\documentclass[review]{elsarticle}

\usepackage{times}
\usepackage{epsfig}
\usepackage{graphicx}
\usepackage{amsmath}
\usepackage{amssymb}
\usepackage{xfrac}
\usepackage{nicefrac}
\usepackage{multirow}
\usepackage{rotating}
\usepackage{tabularx}
\usepackage{array}

\usepackage{amsfonts}
\usepackage{algorithmic}
\usepackage{textcomp}
\usepackage{comment}
\usepackage{arydshln}

\usepackage{hyperref}
\usepackage{times}
\usepackage{epsfig}
\usepackage{booktabs}
\usepackage{algorithm}
\usepackage{paralist}

\newcolumntype{x}[1]{%
>{\centering\hspace{0pt}}p{#1}}%

\newcommand{\figref}[1]{Fig.~\ref{fig:#1}}
\newcommand{\tabref}[1]{Table~\ref{tab:#1}}

\newcommand{\eqnref}[1]{Eq.~(\ref{eqn:#1})}

\newcommand{\etal}{\textit{et al.}}

\usepackage{hyperref}

\journal{Elsevier Image and Vision Computing Journal}

\usepackage{fancyhdr}
\begin{document}
\thispagestyle{fancy}  

\begin{frontmatter}

\title{Loss-Aware Automatic Selection of Structured Pruning Criteria for Deep Neural Network Acceleration}

\author[mymainaddress]{Deepak Ghimire}
\author[mysecondaryaddress]{Kilho Lee}
\author[mymainaddress]{Seong-heum Kim\corref{mycorrespondingauthor}}
\address[mymainaddress]{Visual Intelligence and Platform Lab, College of Information Technology, Soongsil University, Seoul, South Korea}
\address[mysecondaryaddress]{Mobility Intelligence and Computing Systems Lab, College of Information Technology, Soongsil University, Seoul, South Korea}
\cortext[mycorrespondingauthor]{Corresponding author (E-mail: seongheum@ssu.ac.kr)}
\fntext[mymainaddress]{The first and last authors as with the Mobility Intelligence and Computing Systems Lab, College of Information Technology, Soongsil University, Seoul, South Korea.}
\fntext[mysecondaryaddress]{The second author is with the Mobility Intelligence and Computing Systems Lab, College of Information Technology, Soongsil University, Seoul, South Korea.}

%
%

\begin{abstract}
Structured pruning is a well-established technique for compressing neural networks, making it suitable for deployment in resource-limited edge devices. This paper presents an efficient Loss-Aware Automatic Selection of Structured Pruning Criteria (LAASP) for slimming and accelerating deep neural networks. The majority of pruning methodologies employ a sequential process consisting of three stages, 1) training, 2) pruning, and 3) finetuning, whereas the proposed pruning technique adopts a pruning-while-training approach that eliminates the first stage and integrates the second and third stages into a single cycle. The automatic selection of magnitude or similarity-based filter pruning criteria from a specified pool of criteria and the specific pruning layer at each pruning iteration is guided by the network's overall loss on a small subset of the training data. To mitigate the abrupt accuracy drop due to pruning, the network is retrained briefly after each reduction of a predefined number of float point operations (FLOPs). The optimal pruning rates for each layer in the network are automatically determined, eliminating the need for manual allocation of fixed or variable pruning rates for each layer. Experiments on the VGGNet and ResNet models on the CIFAR-10 and ImageNet benchmark datasets demonstrate the effectiveness of the proposed method. In particular, the ResNet56 and ResNet110 models on the CIFAR-10 dataset significantly improve the top-1 accuracy compared to state-of-the-art methods while reducing the network FLOPs by 52\%. Furthermore, the ResNet50 model on the ImageNet dataset reduces FLOPs by more than 42\% with a negligible 0.33\% drop in top-5 accuracy. The source code of this paper is publicly available online. \url{https://github.com/ghimiredhikura/laasp}.
\end{abstract}


\end{frontmatter}

\section{Introduction}
{Convolution} Neural Networks (CNNs) have achieved exceptional performance in a multitude of applications, but this is accompanied by large model sizes and computation requirements, also referred to as FLOPs. To enable the deployment of these complex models on resource-constrained edge devices, it is necessary to reduce memory requirements and computation costs. Model compression, including techniques such as pruning~\cite{han2015deep, han2015learning, frankle2018lottery, ding2019global, he2018soft, he2020learning, he2019filter, he2022filter}, quantization~\cite{jung2019learning, banner2019post, jacob2018quantization}, knowledge distillation~\cite{polino2018model, ji2021refine}, and tensor decomposition~\cite{yang2020learning, yin2021towards} is a popular approach to achieve this goal. Other commonly used solutions include, neural architecture search~\cite{cha2022supernet}, hardware optimization~\cite{chen2019eyeriss}, and hardware-software co-design~\cite{deng2020model, ghimire2022survey}.

Pruning is a promising method for accelerating CNNs and can be classified into weight pruning~\cite{frankle2018lottery, han2015deep, han2015learning, ding2019global} and filter pruning~\cite{li2016pruning, he2018soft, he2022filter, luo2017thinet, yu2018nisp, he2020learning, kim2022automated, mondal2022adaptive}. Weight pruning reduces model size significantly, resulting in a non-structured sparse model that requires specialized hardware to realize its full potential. In contrast, filter pruning eliminates unimportant filters, resulting in a slim yet structured model that is still suitable for general-purpose hardware. 

The early research on weight pruning primarily utilized magnitude-based thresholding to prune weights. In~\cite{han2015deep}, a constant global magnitude threshold was applied, while~\cite{ding2019global} adopted a global compression ratio to compute the per-layer sparsity ratio for weight pruning. Another well-known approach, the lottery ticket hypothesis~\cite{frankle2018lottery} involves utilizing the same initial weights for weight initialization in sub-networks, referred to as "winning tickets," derived from the original network.  

Filter pruning is a valuable technique for compressing CNNs for the purpose of lightweight deep learning. The key advantage of this method is that it results in a structured network that can be executed efficiently on general-purpose hardware, as opposed to a non-structured sparse network, which requires specialized hardware. However, removing the entire filter or channel from the network layers can negatively impact the accuracy of the model, especially when a high pruning rate is applied. Therefore, it is important to carefully select filters and establish a strategy for pruning that compensates for the loss in accuracy.

\begin{figure}[t]
  \centering
  \includegraphics[width=1.0\linewidth]{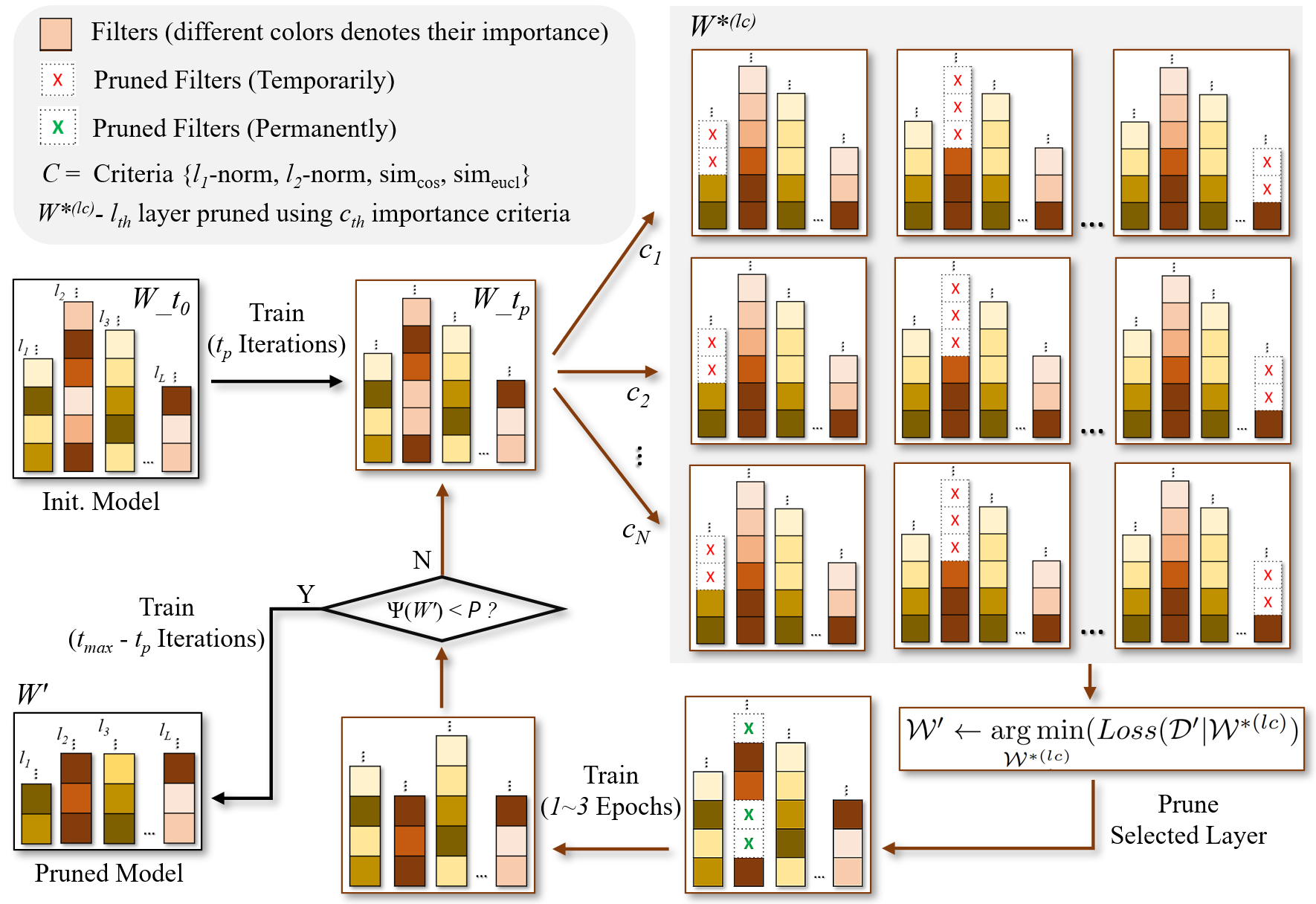}
  \caption{\label{fig:figure1}
  An illustration of the proposed model training and pruning process. Initially, the original model is trained for $t_p$ epochs. At this point, the training is temporarily halted, and the proposed pruning algorithm is applied to this partially trained model to achieve the target pruning rate. Finally, the pruned model is further trained for $t_{max}-t_p$ epochs to converge.}
\end{figure}

Our proposed pruning algorithm is a simple yet effective method of reducing the number of FLOPs in the network. The selection of pruning layers and filter criteria are based on minimizing network loss through a greedy search approach. The proposed algorithm employs a pruning-while-training approach, eliminating the need for an elaborate fine-tuning process. Furthermore, we adopt a hard filter pruning strategy, whereby filters selected for pruning are permanently removed during the iterative pruning process to further decrease network FLOPs. \figref{figure1} illustrates the processing flow of the proposed algorithm.

In summary, the main contributions of this paper are as follows:

\begin{compactitem} 
   \item In the proposed pruning methodology, the traditional train-prune-finetune process is replaced with a pruning-while-training approach, which significantly simplifies the overall procedure. 
   \item In our proposed pruning process, the selection of filters to be pruned in each iteration is automated and guided by the filer importance criteria and data, leading to the optimization of pruning rates for each layer. 
   \item The pruning criteria are selected from the given pool in an automated and data-driven manner, ensuring that the network loss is minimized in each step of the pruning process.
   \item The effectiveness of the proposed pruning algorithm is validated with extensive experiments conducted on CIFAR-10 and ImageNet datasets, utilizing the VGGNet and ResNet architectures. The results of the proposed algorithm are compared against state-of-the-art pruning methods, with a thorough analysis of the pruning outcomes.
\end{compactitem}

The remainder of this paper is organized as follows. Section~\ref{sec:relatedworks} presents a review of related works in the field of filter pruning. The proposed pruning algorithm and the data-driven filter selection strategy are described in detail in Section~\ref{sec:proposedmethod}. The experimental evaluation, comparison with state-of-the-art methods, analysis of the pruning results, and several ablation studies to find optimal hyperparameters used in the pruning algorithm are provided in Section~\ref{sec:experiment}. Finally, Section~\ref{sec:conclusion} concludes the paper and outlines potential avenues for future research.

\section{Related Works}
\label{sec:relatedworks}
The current research focus in the area of filter pruning addresses two core challenges: (1) determining an appropriate criterion for choosing filters to prune, and (2) determining the optimal timing of pruning within the training process. Simple and widely accepted methods for removing unimportant or redundant filters include those based on magnitude or similarity criteria, such as the norm of the filter itself (as seen in~\cite{li2016pruning}, and ~\cite{he2018soft}) or the similarity between filters within the same layer (as seen in~\cite{zuo2020filter}, and ~\cite{he2019filter}). Some studies (such as~\cite{ ghimire2023magnitude}) utilize both of these criteria for filter selection for pruning. 

Filter pruning based on $l_1$-norm is a widely used method for removing filters with low norm values, as seen in studies~\cite{li2016pruning}, and ~\cite{he2018soft}. Recently, Fang~\etal~\cite{fang2023depgraph} first finds all the structural pruning groups and estimates the simple magnitude importance of grouped parameters after learning the group structural sparsity to force coupled parameters to zero. After sparse learning, all the group parameters in a pruning group are used for estimating the group's importance. The FPGM~\cite{he2019filter} algorithm implements redundant filter pruning based on Euclidean similarity criteria, while~\cite{he2022filter} determines the most appropriate pruning criterion based on the network's meta-attributes. However, it can be challenging to determine which filter selection criteria perform optimally for different network architectures. Soft filter pruning techniques, such as those proposed by He~\etal~in \cite{he2018soft},~\cite{he2022filter} and FPGM~\cite{he2019filter} prune filters at the end of each training epoch while still updating them in subsequent epochs, with actual pruning being performed at the end of the training process. In contrast, pruning methods such as ThiNet~\cite{luo2017thinet}, NISP~\cite{yu2018nisp}, and ~\cite{kim2022automated} formulate filter pruning as a complex optimization problem. Jiang~\etal~\cite{di2022channel} proposed to build a graph model for each layer based on the similarity of channel features extracted as global average pooling from a batch of samples. The reinforcement learning is then used to train the set of agents for individual CNN layers to determine the pruning scheme for the overall CNN model. Recently, variable rate filter pruning, where each layer is pruned at a different rate, has gained popularity (as seen in GFI-AP~\cite{mondal2022adaptive},~\cite{kim2022automated, li2022optimizing}). The utilization of the global importance metric is another straightforward technique to prune the filter with adaptive pruning rates in each layer~\cite{fang2023depgraph, mondal2022adaptive}. The pruning techniques discussed above experiment mainly with image classification tasks, whereas the methods for pruning the multitask CNNs~\cite{singh2020falf, ye2023performance} are also available.

The standard method for pruning neural networks involves training the network to its convergence, pruning it based on a predetermined criterion, and then retraining the pruned network to regain accuracy. However, this approach is time-consuming as the network needs to be trained twice. There are also pruning algorithms that can be applied at network initialization, such as the SNIP algorithm proposed by Lee~\etal~\cite{yu2018nisp} and the lottery ticket hypothesis (LTH)~\cite{frankle2018lottery} identify a trainable subnetwork at initialization. Recently, Frankle~\etal~\cite{frankle2021pruning} demonstrated that pruning at initialization can achieve comparable results to magnitude-based pruning of a fully trained network. However, there are some limitations to this approach, such as the risk of suboptimal network structures, subjectivity in choosing the pruning criteria and threshold, and a potential performance gap~\cite{frankle2020linear}. To address these challenges, the pruning-while-training method has emerged as an alternative in which pruning and training are performed in just one training cycle. This approach is similar to the train-prune-retrain scenario, but the pruning is performed on a partially trained original network. Methods such as~\cite{he2019filter, lym2019prunetrain, oyedotun2020structured} allow pruning without requiring a fully trained network. OTOv1~\cite{NEURIPS2021_a376033f} and OTOv2~\cite{chen2023otov2} proposed by Chen~\etal~partition the parameters into zero invariant groups and formulate the structured-sparsity optimization problem to prune zero groups on the only-train-once pipeline. Shen~\etal~\cite{shen2022prune} investigated the required number of initial training epochs for pruning and proposed an early pruning indicator (EPI) to determine the appropriate pruning epoch during network training. In this paper, we adopt a similar approach, but through empirical analysis, we define the early pruning epoch. Our empirical study indicates that a few initial training epochs are sufficient for pruning, followed by training the pruned network for the remaining epochs, resulting in a similar performance compared to the original network. 

\section{Proposed Method}
\label{sec:proposedmethod}
The algorithms utilized in the proposed pruning method are presented in this section. At each stage of pruning, the quantity of filters that will be trimmed is contingent upon the rate of FLOPs reduction. Historically, filter pruning has been carried out based on either magnitude-based criteria or similarity-based criteria. The proposed algorithm, however, dynamically selects the most appropriate criteria from a set of filter selection criteria to ensure optimal pruning. Upon completion of the pruning process, the algorithm returns the pruned model with adaptive pruning rates for each layer.

\subsection{Preliminaries}

Let us assume a neural network has $L$ layers. $N^l_{in}$ and $N^l_{out}$ is the number of input and output channels in the $l_{th}$ convolution layer, respectively. Suppose $\mathcal{F}^l_k\in{\mathbb{R}^{N^l_{in}\times{K^l}\times{K^l}}}$ is the $k_{th}$ filter in the $l_{th}$ convolution layer, where $K^l$ is the kernel size of the filter. Therefore, for the $l_{th}$ layer, network consists set of filters denoted as $\mathcal{F}^l_k, 1\leq k \leq N^l_{out}$ and parameterized by $\mathcal{W}^{(l)} \in {\mathbb{R}^{N^l_{out}\times{N^l_{in}}\times{K^l}\times{K^l}}}, 1 \leq l \leq L$. For simplicity, let us represent all 3-D filters $\mathcal{F}^l_k$ in $l_{th}$ convolution layer as one dimensional vector $\mathcal{X}^l \in \mathbb{R}^{N^l_{out}\times{M^l}}$, which means in $l_{th}$ convolution layer there are $N^l_{out}$ filter vectors and each vector have length $M^l = N^l_{in} \times K^l \times K^l$. 

\subsection{Pruning Objective}

Given a training dataset $\mathcal{D} = \{(a_i,b_i)\}^S_{i=1}$, where $S$ is the total number of training data, $a_i$ and $b_i$ denotes the $i_{th}$ input image and its ground-truth label, respectively. The problem of network pruning with given constraint $\mathcal{P}$ can be formulated as the following optimization problem: 

\begin{equation}
\underset{\mathcal{W}^{\prime}}{\arg\min} \Bigl( Loss(\mathcal{W}^{\prime}, \mathcal{D})\Bigl) ~~~~~\mbox{s.t.}~~~\Psi\Bigl(f(\mathcal{W}^{\prime}, a_i)\Bigl) \leq \mathcal{P} 
\label{eqn:pruneoptfn}
\end{equation}
where $\mathcal{W}^{\prime} \subset \mathcal{W}$ are the parameters of the network remaining after structured pruning. $Loss(.)$ is the network loss, $f(.)$ encodes the network function, and $\Psi(.)$ maps the network to the constraint $\mathcal{P}$, i.e, FLOPs of the network in our case. The method can easily be scaled to other constraints such as latency, or memory. For simplicity, the FLOPs of network parameters $\mathcal{W}$ throughout the paper will be denoted as $\Psi(\mathcal{W})$. The pruning rate of pruned network $\mathcal{W}^{\prime}$ w.r.t. original network $\mathcal{W}$ denoted as $\mathcal{P}^{\prime}$ can be defined as

\begin{equation}
    \mathcal{P}^{\prime}  = 1 - \frac {\Psi(\mathcal{W})} {\Psi(\mathcal{W}^{\prime})}  
  \label{eqn:flopred_rate}
\end{equation}

\subsection{Filter Importance Criterion}

The selection of an appropriate filter importance estimation criterion for ranking filters during pruning is a challenging task, as there is no widely accepted rule among researchers. The distribution of parameters across different pruning groups is influenced by several factors, including the network architecture and the task at hand. To address this challenge, we propose an algorithm that adaptively selects the optimal importance estimation criterion from a pool of available criteria. To minimize the pruning cost, we use simple magnitude and similarity-based criteria, although more complex criteria can also be utilized.

\subsubsection{ Magnitude Criteria}
In the domain of weight pruning, unstructured pruning, the utilization of filter magnitude calculated via norm is a widely adopted criterion. Also, in the context of filter pruning, norm-based measures have been demonstrated to effectively determine the significance of filters. Filters with low norm values are considered to have lower significance as compared to filters with high norm values~\cite{li2016pruning}. If we represent $k_{th}$ filter in $l_{th}$ layer $\mathcal{X}^l_k$ as $\boldsymbol{x} \in \mathbb{R}^{1 \times M_l}$, its $l_p$-norm is calculated as  

\begin{equation}
  {\lVert \mathcal{X}^l_k \lVert}_p = \Biggl({{\sum_{m=1}^{M_l}} {|x_m|}^p }\Biggl)^{1/p}
  \label{eqn:lpnorm}
\end{equation}

In order to incorporate the unique properties of both $l_1$-norm and $l_2$-norm, both measures are included in our pruning algorithm. This approach is taken because the selection of the appropriate norm is performed automatically as part of the pruning process.

\subsubsection{Similarity Criteria}
Filter magnitude calculated using norm-based measures has proven to be effective in identifying and removing unimportant filters. However, it may not always effectively address the issue of redundant filters. To address this, filter similarity measures such as Euclidean similarity and Cosine similarity can be employed. As discussed in~\cite{he2022filter}, these similarity measures can effectively identify and remove redundant filters. 

\textbf{Euclidean Similarity}. 
The Euclidean distance depends upon the magnitude of two vectors. The Euclidean similarity between two filter vectors $\boldsymbol{x} \in \mathbb{R}^{1 \times M_l}$ and  $\boldsymbol{y} \in \mathbb{R}^{1 \times M_l}$ is calculated as 

\begin{equation}
    D_{eucl}(\boldsymbol{x}, \boldsymbol{y}) = \sqrt{ {\sum_{m=1}^{M_l}} {|x_m - y_m|}^2}
  \label{eqn:disteucl}
\end{equation}

\textbf{Cosign Similarity}. 
Cosign similarity measures the cosign of angles between two vectors to determine if they are pointing in the same direction. The Cosign similarity between two filter vectors $\boldsymbol{x} \in \mathbb{R}^{1 \times M_l}$ and  $\boldsymbol{y} \in \mathbb{R}^{1 \times M_l}$ is calculated as 

\begin{equation}
    D_{cos}(\boldsymbol{x}, \boldsymbol{y}) = 1 - \frac {\sum_{m=1}^{M_l} \bigl({x_m \times y_m}\bigl) } { {\sum_{m=1}^{M_l} {x_m^2}} \times {\sum_{m=1}^{M_l} {y_m^2}}}
  \label{eqn:distcos}
\end{equation}

The similarity between two filters in the layer can be quantified using \eqnref{disteucl} and \eqnref{distcos}. However, to determine the overall similarity score for a filter with respect to all other filters in the layer, the final similarity score is computed as the average of the similarity values between the target filter and all other filters in the layer. This average similarity score is used to assess whether the removal of a filter can be compensated by the remaining filters in the layer. The similarity score for a $k_{th}$ filter in a $l_{th}$ convolutional layer is now formulated as

\begin{equation}
    S\bigl(\mathcal{X}^l_k\bigl) = \frac {\sum_{q=1, q \ne k}^{N^l_{out}} D\bigl(\mathcal{X}^l_k, \mathcal{X}^l_q\bigl)} {N^l_{out}-1}
  \label{eqn:davg}
\end{equation}

The calculation of Euclidean and Cosine similarity criteria is performed by substituting $D(\mathcal{X}^l_k, \mathcal{X}^l_q)$ with the expressions provided in~\eqnref{disteucl} and~\eqnref{distcos} respectively. For example, the expression for Cosine similarity score for $k_{th}$ filter in the $l_{th}$ convolution layer becomes

\begin{equation}
    S_{cos}\bigl(\mathcal{X}^l_k\bigl) = \frac {\sum_{q=1, q \ne k}^{N^l_{out}} D_{cos}\bigl(\mathcal{X}^l_k, \mathcal{X}^l_q\bigl)} {N^l_{out}-1}
  \label{eqn:davgex}
\end{equation}

\subsection{Layer-wise Exploration Step Estimation}

The removal of a filter from a convolutional layer has far-reaching impacts, not only affecting the preceding layer and the subsequent layer but also on other parts of the network if there are other connections such as shortcut connections. Since the 3-D shape of each layer is unique, the reduction in FLOPs will vary even if the same number of filters are pruned from different layers. To minimize the network loss and ensure a fair comparison among layers, our method seeks to determine a suitable number of filters to prune for each layer in each iteration so that the reduction in network FLOPs is almost constant. In this paper, we named it as a layer-wise exploration step. Let us suppose $\mathcal{W}^{\bar{(l)}}$ denotes the network parameters excluding $l_{th}$ layer parameters. Now, if we fix the network FLOPs reduction rate per exploration to $\mathcal{P}_s$ throughout the pruning process, the exploration step for $l^{th}$ layer denoted by ${E}^{(l)}_s$ can be approximated by the following equation

\begin{equation}
    E^{(l)}_s = \max \Biggl(1, \frac {\mathcal{P}_s \times {\Psi(\mathcal{W})} \times {N_{l+1}} } { \bigl(\Psi(\mathcal{W}) - \Psi(\mathcal{W}^{\bar{(l)}}\bigl) } \Biggl) 
  \label{eqn:step_rate}
\end{equation}

\begin{figure}[thpb]
  \centering
  \includegraphics[width=1.0\linewidth]{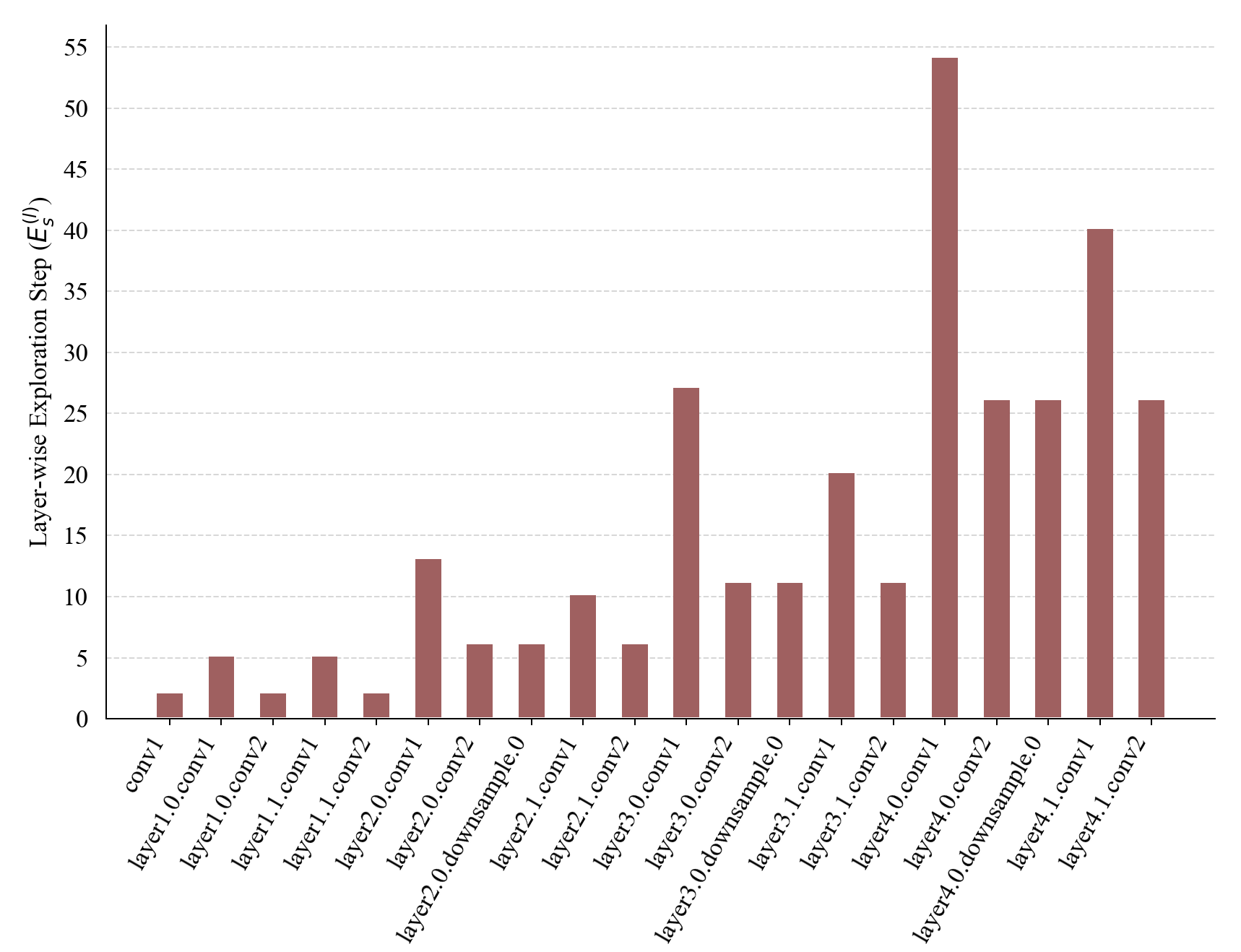}
  \caption{Step pruning rate for each layer of the ResNet18 model evaluated in ImageNet dataset. The overall FLOPs reduction rate was fixed at approximately $1\%$ in this experiment.}
  \label{fig:figure2}
\end{figure}

\figref{figure2} depicts the layer-wise exploration rate of the filters in the ResNet18 model, as evaluated on the ImageNet dataset, using~\eqnref{step_rate} to ensure a FLOPs reduction rate of approximately $1\%$ in each pruning iteration, i.e., also the exploration step. For instance, as shown in~\figref{figure2}, from the $conv1$ convolution layer, we need to remove two filters to achieve a $1\%$ reduction in overall network FLOPs, while the $layer1.0.conv1$ convolution layer necessitates the pruning of five filters to reach the same reduction rate. The exploration rate in each iteration, as depicted in this paper, is set to approximate a $1\%$ FLOPs reduction, though it can be adjusted to accommodate different FLOPs reduction rates as desired by varying the value of $\mathcal{P}_s$ in~\eqnref{step_rate}. To ensure minimal complexity during the pruning process, we calculated the exploration step based on the original network at the beginning and maintained it throughout the pruning iterations. However, it is important to note that the exploration step may change as we continue to advance through the pruning iterations.

\subsection{Pruning Algorithm}

The presented methodology circumvents the requirement of a completely trained pre-trained network for pruning, by conducting pruning on a partially trained network over several training epochs. Upon reaching the targeted pruning rate, training is resumed until convergence to obtain the final pruned network. At each pruning iteration, the pruning layer along with pruning criteria are selected in such a way that the overall network loss is minimal due to pruning. We use a greedy search approach to find the optimal set of pruning filters at each pruning iteration which results in variable rate pruning. In each pruning iteration, the selection of a subset of filters from the entire network is a two-step process. Firstly, we rank the filters based on all importance estimation criteria for each pruning layer. Secondly, we calculate the network loss by removing the low-ranking filters in each layer for all criteria. From all available options, we select the candidate for pruning that has the least amount of loss. As it is necessary to calculate the network loss for multiple subnetworks, we use a small, randomly sampled training dataset for loss estimation in order to minimize computation costs.

\begin{algorithm}
 \caption{Algorithm of LAASP}
 \begin{algorithmic}[1] \label{alg:algorithm1}
 \renewcommand{\algorithmicrequire}{\textbf{Input:}}
 \renewcommand{\algorithmicensure}{\textbf{Output:}}
 \REQUIRE training data $(\mathcal{D})$, initial model ($\mathcal{W}$) 
 \STATE \textbf{Given:} number of training epochs before pruning ($t_p$), target FLOPs reduction rate ($\mathcal{P}$), layer-wise max pruning rate (${R}^{(l)}_{max}$), exploration step FLOPs reduction rate ($\mathcal{P}_s$), and fine-tune FLOPs reduction interval ($\delta\mathcal{P}_{ft}$)
 \STATE \textbf{Initialization:} $\mathcal{P}^{\prime} \gets 0$, $\mathcal{P}^{\prime}_{temp} \gets 0$
 \STATE Randomly sample dataset $\mathcal{D}^{\prime}$ from $\mathcal{D}$
 \STATE Train original model $\mathcal{W}$ for $t_p$ epochs ({$epoch \gets 0$ to $t_p$}) \label{lst:line:inittrain}
 \FOR {$l \gets 1$ to $L$} \label{lst:line:step_rate_start}
 \STATE Calculate step pruning rate ($E^{(l)}_s$) using~\eqnref{step_rate}
 \ENDFOR \label{lst:line:step_rate_end}
 \STATE Set $\mathcal{W}^{\prime} \gets \mathcal{W}$
 \WHILE{$\mathcal{P}^{\prime} < \mathcal{P}$} \label{lst:line:targetflop}
  \FOR {$l \gets 1$ to $L$} \label{lst:line:scanlayers}
  \IF { $ N^{{\prime}l}_{out} \geq N^l_{out} \times {R}^{(l)}_{max}$ } \label{lst:line:pruninglimit}
  \FOR {$c \gets 1$ to $C$} \label{lst:line:pruningcriteria}
  \STATE Calculate rank of $l_{th}$ layer filters using $c_{th}$ filter selection criteria (~\eqnref{lpnorm} or ~\eqnref{davg}) 
  \STATE $\mathcal{W}^{{*(lc)}} \gets$ prune $E^{(l)}_s$ lowest rank filters  \label{lst:line:prunecandidate}
  \ENDFOR
  \ENDIF
  \ENDFOR \label{lst:line:magpruneend}
  \STATE $\mathcal{W}^{\prime} \gets \underset{\mathcal{W}^{*(lc)}}{\arg\min} ( Loss(\mathcal{D}^{\prime}|\mathcal{W}^{*(lc)}) $ \label{lst:line:smallloss}
  \STATE Calculate $\mathcal{P}^{\prime}$ using~\eqnref{flopred_rate}
  \IF {$\mathcal{P}^{\prime} - \mathcal{P}^{\prime}_{temp} \geq \delta\mathcal{P}_{ft}$} \label{lst:line:prunerecover}
  \STATE Fine-tune $\mathcal{W}^{\prime}$ to recover weights ($epoch \gets 1$ to $epoch_{ft}$)
  \STATE $\mathcal{P}^{\prime}_{temp} \gets \mathcal{P}^{\prime}$
  \ENDIF \label{lst:line:prunerecoverend}
 \ENDWHILE
 \STATE Train pruned model $\mathcal{W}^{\prime}$ until convergence ({$epoch \gets t_p+1$ to $t_{max}$}) \label{lst:line:pruneretrain}
 \ENSURE pruned model $(\mathcal{W}^{\prime})$
 \end{algorithmic} 
\end{algorithm}

Given a dataset $\mathcal{D}$, $\mathcal{D}^{\prime}$ is the randomly sampled small subset of this dataset. As we denote $\mathcal{W}^{\prime(l)}$ as the filter of the $l_{th}$ layer after pruning, and $N^{{\prime}l}_{out}$, the number of filters after pruning, the estimation of accuracy loss during each pruning step is denoted as $Loss(\mathcal{D}^{\prime}|\mathcal{W}^{\prime})$. In each pruning iteration, we temporarily prune each layer using each filter selection criteria ($C$ = \{$l_1$-norm, $l_2$-norm, eucl-sim, cos-sim\}) and select the particular layer and pruning criteria that result in minimal accuracy loss. The number of filters to be pruned in each step for each layer is pre-calculated such that the network FLOPs reduction rate remains constant ($\mathcal{P}_s$) using~\eqnref{step_rate}. To mitigate excessive pruning from a single convolutional layer and to solve the probable layer collapse issue, a maximum pruning rate limit of $R^{(l)}_{max}$ is imposed. For instance, if $R^{(l)}_{max}$ is set to $0.7$, the maximum number of filters that can be pruned from the $l_{th}$ convolutional layer is capped at $N^l_{out} \times 0.7$. While it would be valuable to explore the optimal pruning limit for each layer for a target pruning rate, we have opted for simplicity and set a fixed pruning limit for all layers in this paper. 

The detailed pruning steps of the proposed pruning algorithm are outlined in Algorithm~\ref{alg:algorithm1}. Given the training dataset ($\mathcal{D}$), the algorithm begins by training the original network for a specified number of epochs ($t_p$) (Line~\ref{lst:line:inittrain}), before pausing the training process. The step pruning rate is then calculated for each layer of the network (Line~\ref{lst:line:step_rate_start}-\ref{lst:line:step_rate_end}) to ensure that a predefined percentage of FLOPs ($\mathcal{P}_s$) is reduced in each pruning iteration. Subsequently, an iterative pruning algorithm is used to achieve the target FLOPs reduction rate by pruning a predetermined number of filters (exploration step) in the selected layer in each pruning iteration (Line~\ref{lst:line:targetflop}). In each iteration, the algorithm scans through every layer (Line~\ref{lst:line:scanlayers}) and pruning criterion ($C$) (Line~\ref{lst:line:pruningcriteria}) to temporarily prune a set of filters (${E}^{(l)}_s$) and generate the set of candidate pruned models $\mathcal{W}^{{*(lc)}}$ (Line~\ref{lst:line:prunecandidate}). To avoid excessive pruning from a single convolution layer, the maximum number of pruned filters in each layer is limited to $N^l_{out} \times R^{(l)}_{max}$ (Line~\ref{lst:line:pruninglimit}). The pruned model that results in the minimal accuracy loss among the candidates is then selected for the current iteration (Line~\ref{lst:line:smallloss}) and the pruning is made permanent. To prevent a significant drop in accuracy during pruning, the pruned model is fine-tuned for a small number of epochs ($epoch_{ft}$) after reducing $\delta\mathcal{P}_{ft}$ FLOPs (Line~\ref{lst:line:prunerecover}). Once the target pruning rate is reached, the original training is resumed and the pruned model is trained until convergence ($t_{max} - t_p$ epochs) to obtain the final fully trained pruned model $\mathcal{W}^{\prime}$ (Line~\ref{lst:line:pruneretrain}).

\section{Experimental Results}
\label{sec:experiment}

\subsection{Experimental Settings}

To evaluate the effectiveness of our pruning technique, we conducted experiments on two widely used datasets, namely CIFAR-10~\cite{krizhevsky2009learning} and ImageNet~\cite{russakovsky2015imagenet}. The CIFAR-10 dataset comprises 50,000 training images and 10,000 validation images, organized into 10 classes. On the other hand, the ImageNet dataset encompasses 1.28 million training images and 50,000 validation images, distributed across 1,000 classes. Our primary focus was on pruning multi-branch residual network architectures (ResNet~\cite{he2016deep}) and we evaluated the performance on both CIFAR-10 and ImageNet datasets. In addition, we also examined the pruning of the single-branch VGGNet~\cite{simonyan2014very}, which is relatively simple, and evaluated its performance only on the CIFAR-10 dataset.

When training ResNet and VGGNet on the CIFAR-10 dataset, we utilized the same training settings as reported in~\cite{he2016identity, zagoruyko2016wide} and~\cite{li2016pruning, he2019filter}, respectively. For training ResNet on the ImageNet dataset, we employed the default parameter settings from~\cite{he2016deep} and adopted the data augmentation strategies from the official PyTorch~\cite{paszke2017automatic} examples. Our proposed algorithm follows a pruning-while-training approach, where the original network is initially trained for a certain number of epochs and then the training process is paused to perform the pruning. After completing the pruning, the training of the pruned network is resumed with the same training settings used for the original network.  Empirically, we found that pruning the network in the early training epochs, just before the initial learning rate is decayed, results in well-recovered network accuracy after pruning.

We employed the PyTorch deep learning framework~\cite{paszke2017automatic} and the structural pruning tool~\cite{fang2023depgraph} for training and pruning the neural network, respectively. At the end of each pruning iteration, the pruned filters were permanently removed, and the network was reconfigured for the subsequent pruning iteration using~\cite{fang2023depgraph}. The maximum pruning rate for each layer was determined empirically in the range $0.55 \leq {R}^{(l)}_{max} \leq 0.75$. If a higher desired pruning rate was required, a larger ${R}^{(l)}_{max}$ value was set, and if a lower pruning rate was desired, a smaller ${R}^{(l)}_{max}$ value was used. In each pruning iteration, the FLOPs reduction rate was set to approximately $1\%$. During the pruning process, after every $\beta_{\mathcal{P}} = 0.03$ (i.e., $3\%$) reduction in network FLOPs, the pruned network was fine-tuned for a limited number of epochs, ranging from $1$ to $3$.

\subsection{VGGNet on CIFAR-10}

\tabref{tbl1} presents a comparison of the VGG16 pruning outcomes on the CIFAR-10 database with the state-of-the-art methods. Initially, the original network is trained for 30 epochs, and pruning is performed. To match other methods with comparable pruning ratios, the network is first pruned iteratively to reduce FLOPs by $34.6\%$. As evident from \tabref{tbl1}, the accuracy of the pruned network is better than other methods. Additionally, with the proposed pruning technique, the FLOPs of the network can be reduced by over $60\%$ while maintaining the same accuracy as the original unpruned network. PFEC~\cite{li2016pruning}, HRank~\cite{lin2020hrank}, and CPGCN~\cite{di2022channel} used the pretrained network for pruning. Similar to our method, CPGCN~\cite{di2022channel} also utilizes both similarity and magnitude-based importance estimation matrices, but on the channel features extracted as global average pooling from a batch of samples. 

\begin{table} [H]
\caption{Comparison of VGG16 pruning on the CIFAR-10 dataset.}
\label{tab:tbl1}
  \begin{center}
    {\small{
\begin{tabular}{lcllcc}
\toprule
& {Pretrain} & {Baseline} & {Pruned } & {Top-1 acc.} & {FLOPs } \\
{Method} & {Used?} & { top-1 acc. (\%)} & {top-1 acc. (\%)} & {drop (\%)} & {($\downarrow$) (\%)} \\
\midrule
    PFEC~\cite{li2016pruning} & Yes & 93.58 $\pm$ 0.03 & 93.31 $\pm$ 0.02 & 0.27 & 34.2 \\
    MFP~\cite{he2022filter} & No & 93.58 $\pm$ 0.03 & 93.54 $\pm$ 0.03 & {0.04} & 34.2 \\
    FPGM~\cite{he2019filter} & No & 93.58 $\pm$ 0.03 & {93.54} $\pm$ 0.08 & {-0.04} & 34.2 \\ 
    LAASP(ours) & No & \textbf{93.79 $\pm$ 0.23} & \textbf{93.90 $\pm$ 0.16} & \textbf{-0.11} & \textbf{34.6} \\
    \hdashline
    HRank~\cite{lin2020hrank} & Yes & \textbf{93.96} & 93.43 & 0.53 & 53.5 \\
    CPGCN~\cite{di2022channel} & Yes & 93.2 & 93.53 & -0.51 & 57.3 \\
    LAASP(ours) & No & {93.79 $\pm$ 0.23} & \textbf{93.79 $\pm$ 0.11} & \textbf{0.00} & \textbf{60.5} \\
\bottomrule
\end{tabular}
}}
\end{center}
\end{table}

MFP~\cite{he2022filter} and FPGM~\cite{he2019filter} are two techniques that do not require a pre-trained network. Instead, they perform filter ranking and zeroing of low-rank filters during each training epoch, and the actual removal of filters is performed only at the end of the training process. In contrast, our method physically removes filters and their corresponding channel dimensions throughout the network in each pruning step. As we pruned the network after only 30 epochs of training from scratch, we don't have a baseline model to compare the accuracy drop. To address this issue, we also suggest training the original network, which was trained for 30 epochs, until convergence without pruning. We can then use this model as the baseline for accuracy drop comparison. The reported baseline accuracies were obtained through a consistent methodology across all experiments.

\subsection{ResNet on CIFAR-10}

\tabref{tbl2} presents the efficiency of the pruning of ResNet34, ResNet56, and ResNet110 for the CIFAR-10 dataset in comparison with state-of-the-art methods. Our pruning algorithm achieves comparable, and even superior, top-1 accuracy with similar FLOP reduction rates. The multiple entries of the proposed method are for different FLOPs reduction rates. Pruning methods such as SFP~\cite{he2018soft}, FPGM~\cite{he2019filter}, and MPF~\cite{he2022filter} adopt the soft filter pruning technique, where the network structure remains unchanged before and after pruning. In contrast, our approach utilizes hard filter pruning, where filters are permanently removed and network connections are reorganized for further processing. This approach has the advantage of reducing the computation cost because after pruning is completed the training will be performed on the slimmed model for convergence. As we are mainly interested to compare the results of pruning methods where pretraining is not required, in order to show the competitiveness of our methods we also listed pruning results from a few other popular methods (GFI-AP~\cite{mondal2022adaptive}, HRank~\cite{lin2020hrank}, NPPM~\cite{gao2021network}, DepGraph~\cite{fang2023depgraph}, Rethink~\cite{liu2018rethinking}, MSVFP~\cite{ghimire2023magnitude}) which performs pruning on the fully trained network. For DepGraph~\cite{fang2023depgraph}, sparsity learning was not employed in the reported results, as doing so would require additional training expenses and thereby render a fair comparison with our proposed approach unfeasible.

\begin{table}
\caption{Comparison of pruned ResNet models on the CIFAR-10 dataset. {FLOPs RR ($\downarrow$) (\%)} denotes the reduction rate of floating-point operations between the baseline and pruned models. The "Acc. drop" metric represents the discrepancy in top-1 validation accuracy between the baseline and pruned models.}
\label{tab:tbl2}
  \begin{center}
    {\small{
\begin{tabular}{clcllcc}
\toprule
\multicolumn{1}{c}{Depth} & {Method} & {Pretrain Used?} & {Baseline acc. (\%)} & {Pruned acc. (\%)} & {Acc. drop (\%)} & {FLOPs RR ($\downarrow$) (\%)} \\
\midrule
\multirow{5}{*}{32} & SFP~\cite{he2018soft} & No & 92.63 $\pm$ 0.70 & 92.08 $\pm$ 0.08 & 0.55 & 41.5 \\ 
                    & GFI-AP~\cite{mondal2022adaptive} & Yes & 92.54 & 92.09 $\pm$ 0.15 & {0.45} & \textbf{42.5} \\
                    & LAASP(ours) & No & \textbf{93.12 $\pm$ 0.04} & \textbf{92.71 $\pm$ 0.22} & \textbf{0.41} & \textbf{42.5} \\
                    \cdashline{2-7}
                    & FPGM~\cite{he2019filter} & No & 92.63 $\pm$ 0.70 & {91.93 $\pm$ 0.03} & {0.70} & 53.2 \\
                    & MFP~\cite{he2022filter} & No & 92.63 $\pm$ 0.70 & 91.85 $\pm$ 0.09 & 0.78 & 53.2 \\
                    & LAASP(ours) & No & \textbf{93.12 $\pm$ 0.04} & \textbf{92.64 $\pm$ 0.09} & \textbf{0.48} & \textbf{53.3} \\
\midrule
\multirow{7}{*}{56} & PFEC~\cite{li2016pruning} & Yes & 93.04 & 93.06 & {-0.02} & 27.6 \\
                    & HRank~\cite{lin2020hrank} & Yes & 93.26 & 93.17& 0.09 & 50.0 \\
                    & NPPM~\cite{gao2021network} & Yes & 93.04 & 93.40 & \textbf{-0.36} & 50.0 \\
                    & SFP~\cite{he2018soft} & No & 93.59 $\pm$ 0.58 & 92.26 $\pm$ 0.31 & 1.33 & \textbf{52.6} \\ 
                    & FPGM~\cite{he2019filter} & No & 93.59 $\pm$ 0.58 & 92.93 $\pm$ 0.49 & 0.66 & \textbf{52.6} \\
                    & MFP~\cite{he2022filter} & No & 93.59 $\pm$ 0.58 & {92.76 $\pm$ 0.03} & 0.83 & \textbf{52.6} \\
                    & DepGraph~\cite{fang2023depgraph} & Yes & 93.53 & 93.46 & 0.07 & \textbf{52.6} \\
                    & LAASP(ours) & No & \textbf{93.61 $\pm$ 0.11} & \textbf{93.49 $\pm$ 0.00} & 0.12 & \textbf{52.6} \\
                    \cdashline{2-7}
                    & LAASP(ours) & No & \textbf{93.61 $\pm$ 0.11} & \textbf{93.04 $\pm$ 0.08} & 0.57 & \textbf{55.2} \\
\midrule
\multirow{8}{*}{110} & PFEC~\cite{li2016pruning} & Yes & 93.53 & 93.30 & 0.23 & 38.6 \\
                    & SFP~\cite{he2018soft} & No & 93.68 $\pm$ 0.32 & 93.38 $\pm$ 0.30 & {0.30} & 40.8 \\ 
                    & Rethink~\cite{liu2018rethinking} & Yes & 93.77 $\pm$ 0.23 & 93.70 $\pm$ 0.16 & 0.07 & 40.8 \\
                    & FPGM~\cite{he2019filter} & No & 93.68 $\pm$ 0.32 & 93.73 $\pm$ 0.23 & {-0.05} & 52.3 \\
                    & MFP~\cite{he2022filter} & No & 93.68 $\pm$ 0.32 & 93.69 $\pm$ 0.31 & -0.01 & 52.3 \\
                    & MSVFP~\cite{ghimire2023magnitude} & Yes & 93.69 $\pm$ 0.22 & 93.92 $\pm$ 0.52 & \textbf{-0.23} & 52.4 \\
                    & LAASP(ours) & No & \textbf{94.41 $\pm$ 0.07} & \textbf{94.17 $\pm$ 0.16} & 0.24 & \textbf{52.5} \\
                    \cdashline{2-7}
                    & HRank~\cite{lin2020hrank} & Yes & 93.50 & 93.36 & \textbf{0.14} & {58.2} \\
                    & LAASP(ours) & No & \textbf{94.41 $\pm$ 0.07} & \textbf{93.58 $\pm$ 0.21} & 0.83 & \textbf{58.5} \\
\bottomrule
\end{tabular}
}}
\end{center}
\end{table}

\begin{figure} [H]
  \centering
  \includegraphics[width=1.0\linewidth]{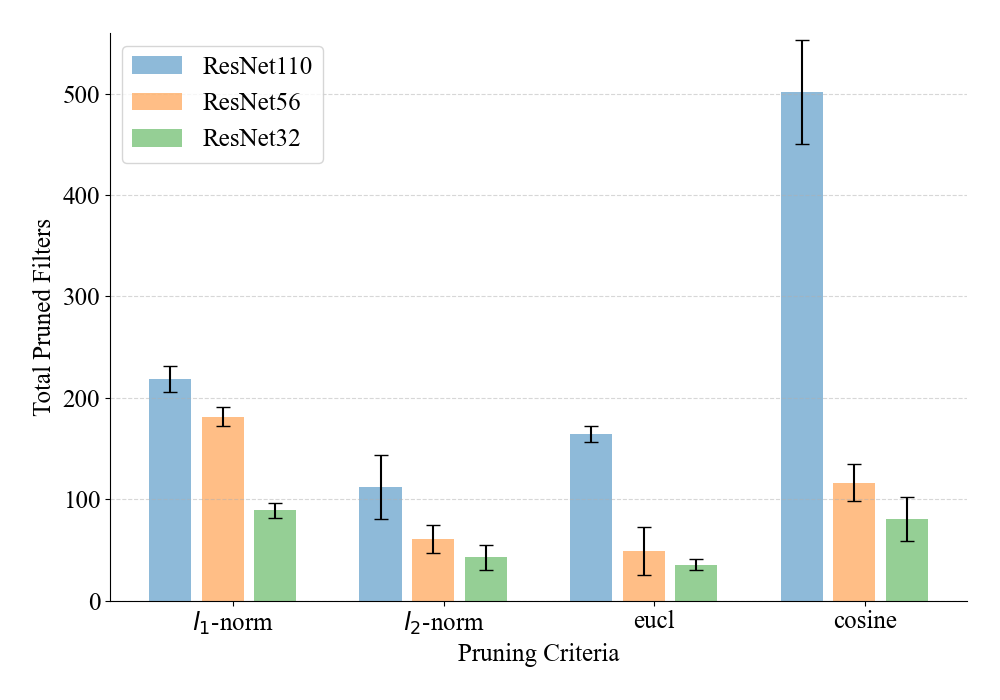}
  \caption{The distribution of the total number of pruned filters for ResNet32, ResNet56, and ResNet110 on the CIFAR-10 dataset among various filter selection criteria while achieving a reduction in network FLOPs of over 50\%.}
  \label{fig:figure3}
\end{figure}

Our algorithm automatically selects the optimal filter importance criteria while pruning filters from each layer.~\figref{figure3} depicts the distribution of pruned filters among four different filter selection criteria from three different runs with error bars. For the ResNet110 model, the majority of filters are removed using the cosine similarity measure, while for the ResNet56 model, most filters are removed using the $l_1$-norm magnitude-based criteria. This suggests that the ResNet110 model has several redundant filters, while the ResNet56 model may have low-magnitude filters. 

\begin{figure} [H]
  \centering
  \includegraphics[width=1.0\linewidth]{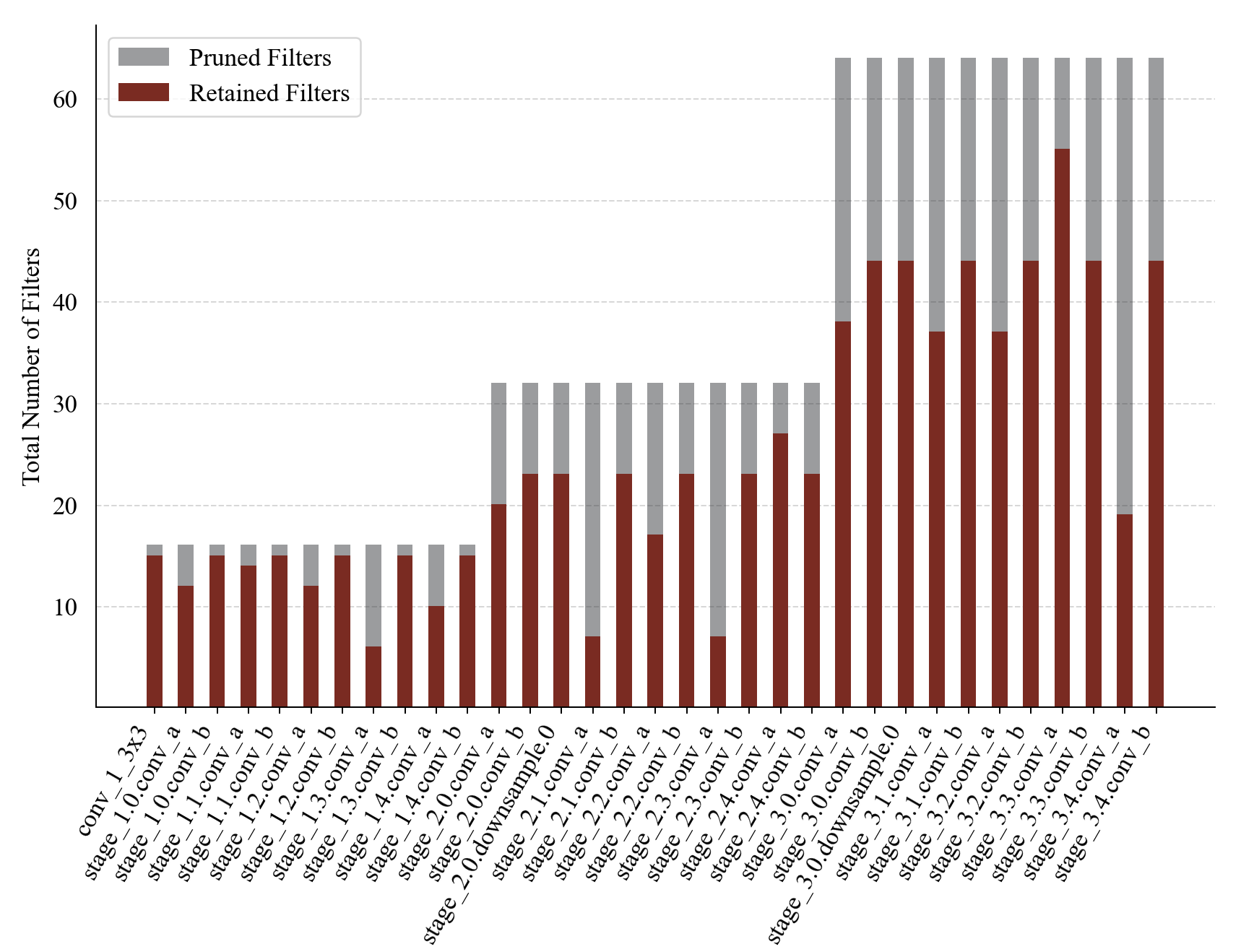}
  \caption{The number of pruned and retained filters in each layer of ResNet32 after pruning on the CIFAR-10 dataset while reducing the network FLOPs by 53.3\%.}
  \label{fig:figure4}
\end{figure}

Finally,~\figref{figure4} shows the ratio of pruned and retained filters in each convolutional layer of the ResNet32 model on the CIFAR-10 dataset. As we can see, in each stage of the ResNet32 model, more filters are removed from the $conv\_a$ layer as compared to the $conv\_b$ layer. This is because pruning a filter in $conv\_b$ layer has a chain effect across channel dimensions of the ResNet models resulting in more FLOPs reduction as compared to pruning a filter from $conv\_a$ layer, which we can also relate with~\figref{figure2}.

\subsection{ResNet on ImageNet}

\begin{table}
\caption{Comparison of pruned ResNet models on ImageNet dataset. The {"FLOPs RR ($\downarrow$) (\%)"} metric indicates the percentage reduction of floating-point operations between the baseline and pruned models. The {"Acc. ($\downarrow$) (\%)"} metric denotes the difference in validation accuracy expressed as a percentage between the baseline and pruned models.}
\label{tab:tbl3}
  \begin{center}
    {\small{
\begin{tabular}{clcccccccc}
\toprule
& & Pretrain & Baseline & Pruned & Top-1 acc. & Baseline & Pruned & Top-5 acc. & FLOPs \\
Depth & Method & Used? & top-1 acc. & top-1 acc. & drop ($\downarrow$) & top-5 acc. & top-5 acc. & drop ($\downarrow$) & RR ($\downarrow$) \\
& & & (\%) & (\%) & (\%) & (\%) & (\%) & (\%) & (\%) \\
\midrule
\multirow{7}{*}{18} & SFP~\cite{he2018soft} & No & 70.28 & 67.10 & 3.18 & \textbf{89.63} & 87.78 & 1.85 & 41.8 \\ 
                    & FPGM~\cite{he2019filter} & No & 70.28 & 67.78 & 2.50 & \textbf{89.63} & 88.01 & 1.62 & 41.8 \\ 
                    & FuPruner~\cite{li2020fusion} & Yes & 69.76 & 68.24 & \textbf{1.52} & 89.08 & 88.21 & \textbf{0.87} & 41.8 \\ 
                    & MFP~\cite{he2022filter} & No & 70.28 & 67.66 & 2.62 & \textbf{89.63} & 87.90 & 1.73 & 41.8 \\ 
                    & LAASP(ours) & No & \textbf{70.58} & \textbf{68.66} & 1.92 & 89.57 & \textbf{88.50} & 1.07 & \textbf{42.2} \\ 
                    \cdashline{2-10}
                    & LAASP(ours) & No & \textbf{70.58} & \textbf{68.12} & \textbf{2.46} & \textbf{89.57} & \textbf{88.07} & \textbf{1.50} & \textbf{45.4} \\
\midrule
\multirow{6}{*}{34} & PaT~\cite{shen2022prune} & No & - & {73.50} & - & - & - & - & {20.7} \\ 
                    & PFEC~\cite{li2016pruning} & Yes & {73.23} & 72.17 & {1.06} & - & - & - & {24.2} \\ 
                    \cdashline{2-10}
                    & SFP~\cite{he2018soft} & No & \textbf{73.92} & 71.83 & 2.09 & 91.62 & 90.33 & 1.29 & 41.1 \\ 
                    & FPGM~\cite{he2019filter} & No & \textbf{73.92} & 71.79 & 2.13 & 91.62 & {90.70} & {0.92} & 41.1 \\ 
                    & FuPruner~\cite{li2020fusion} & Yes & 73.32 & 72.14 & \textbf{1.18} & 91.42 & 90.66 & 0.75 & 41.1 \\ 
                    & LAASP(ours) & No & 73.90 & \textbf{72.65} & 1.25 & \textbf{91.72} & \textbf{90.98} & \textbf{0.74} & \textbf{41.4} \\ 
                    \cdashline{2-10}
                    & LAASP(ours) & No & \textbf{73.90} & \textbf{72.37} & \textbf{1.53} & \textbf{91.72} & \textbf{90.80} & \textbf{0.92} & \textbf{45.4} \\ 
\midrule
\multirow{9}{*}{50} & ThiNet~\cite{luo2017thinet} & Yes & 72.88 & 72.04 & 0.88 & 91.14 & 90.67 & 0.47 & 36.7 \\ 
                    & SFP~\cite{he2018soft} & No & 76.15 & 74.61 & 1.54 & 92.87 & 92.06 & 0.81 & 41.8 \\ 
                    & FPGM~\cite{he2019filter} & No & 76.15 & 75.03 & 1.12 & 92.87 & 92.40 & 0.47 & 42.2 \\ 
                    & LAASP(ours) & No & \textbf{76.48} & \textbf{75.85} & \textbf{0.63} & \textbf{93.14} & \textbf{92.81} & \textbf{0.33} & \textbf{42.3} \\ 
                    \cdashline{2-10}
                    & LFC~\cite{singh2020leveraging} & Yes & 75.30 & 73.40 & 1.90 & 92.20 & 91.40 & 0.80 & 50.0 \\ 
                    & GFI-AP~\cite{mondal2022adaptive} & Yes & 75.95 & 74.07 & 1.88 & - & - & - & 51.9 \\ 
                    & MFP~\cite{he2022filter} & No & 76.15 & 74.13 & {2.02} & 92.87 & {91.94} & {0.93} & 53.5 \\ 
                    & MFP~\cite{he2022filter} & Yes & 76.15 & 74.86 & 1.29 & 92.87 & 92.43 & \textbf{0.44} & 53.5 \\ 
                    & LAASP(ours) & No & \textbf{76.48} & \textbf{75.44} & \textbf{1.04} & \textbf{93.14} & \textbf{92.59} & 0.55 & \textbf{53.9} \\
                    \cdashline{2-10}
                    & PaT~\cite{shen2022prune} & No & - & 74.85 & - & - & - & - & \textbf{58.7} \\ 
\bottomrule
\end{tabular}
}}
\end{center}
\end{table}

The effectiveness of our pruning method was evaluated on the ILSVRC-2012 (ImageNet) dataset for the ResNet-18, ResNet-34, and ResNet-50 models. Each model was tested with two different FLOPs reduction rates, and a specific rate was selected to align with commonly used pruning rates by other researchers. The results of our method are compared with various state-of-the-art pruning methods such as PFEC~\cite{li2016pruning}, SFP~\cite{he2018soft}, FPGM~\cite{he2019filter}, MFP~\cite{he2022filter}, GFI-AP~\cite{mondal2022adaptive}, ThiNet~\cite{luo2017thinet}, FuPruner~\cite{li2020fusion}, PaT~\cite{shen2022prune} in~\tabref{tbl3}, when pruning the  ResNet on the ImageNet dataset. For all models, our method achieved the highest top-1 accuracy among the other methods with similar pruning rates. Particularly, for the ResNet-50 model, although our baseline accuracy was the highest among others, with a $42.3\%$ FLOPs reduction rate, there was only a $0.33\%$ drop in top-5 accuracy. Consistently superior outcomes are observed with the proposed methods as compared to those employing pretrained networks or not utilizing pretrained networks. In the realm of pruning techniques, the PaT method, as described in the paper by Shen~\etal~\cite{li2020fusion}, bears the most similarity to our approach in terms of early pruning during training. The PaT method seeks to determine the optimal early pruning epoch by closely monitoring the stability of the network throughout the initial stages of training. In contrast, our technique involves utilizing a fixed early epoch for pruning that has been identified through empirical analysis. 

\begin{figure}[H]
  \centering
  \includegraphics[width=1.0\linewidth]{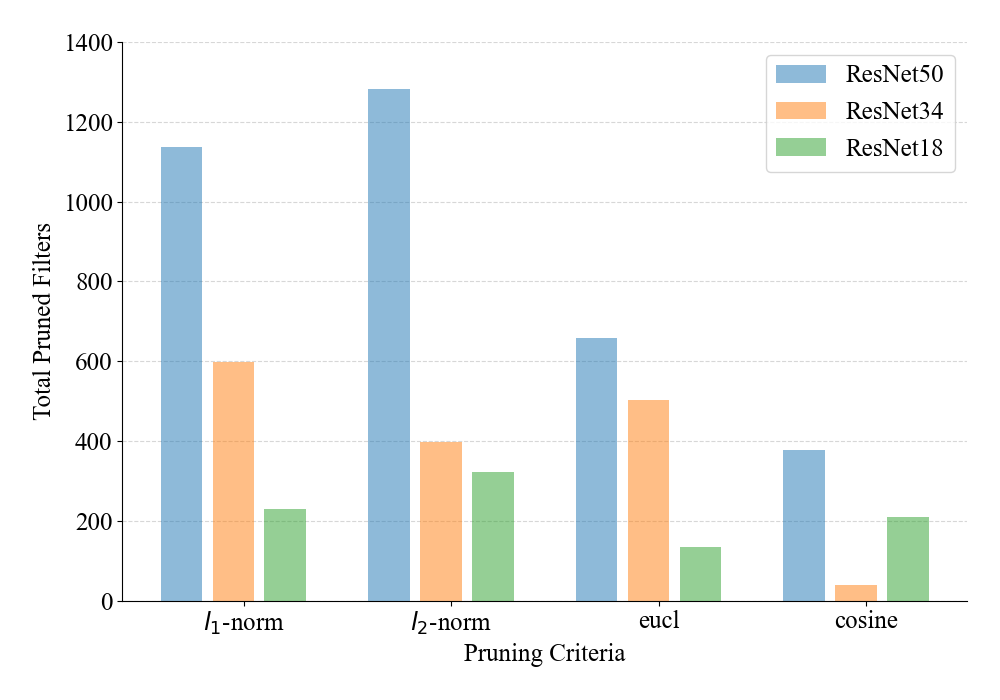}
  \caption{The distribution of the total number of pruned filters for ResNet18, ResNet34, and ResNet50 on the ImageNet dataset among various filter selection criteria, resulting in a reduction of network FLOPs by $45\%$, $45\%$, and $54\%$ respectively.}
  \label{fig:figure5}
\end{figure}

\begin{figure}[H]
  \centering
  \includegraphics[width=1.0\linewidth]{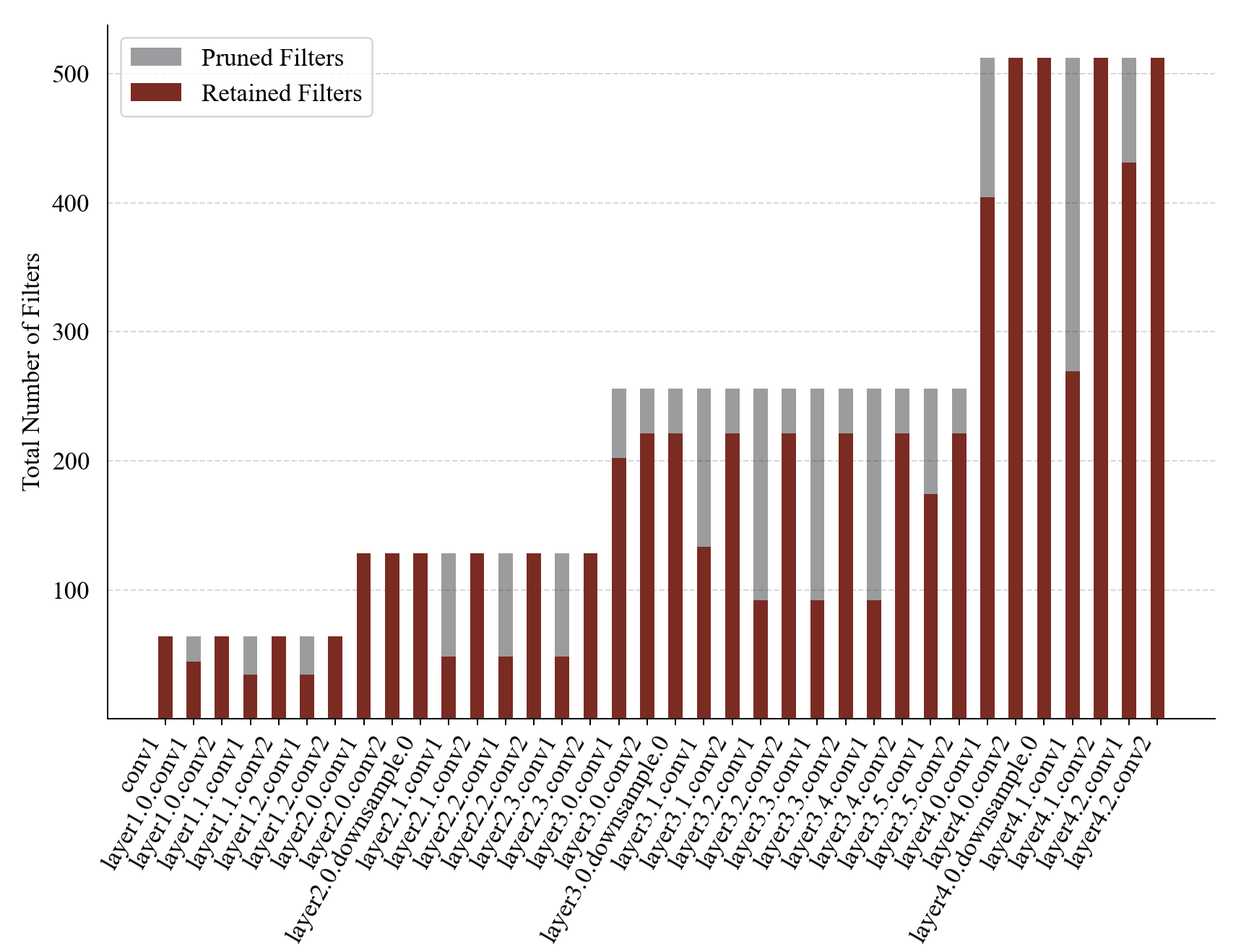}
  \caption{The number of pruned and retained filters in each layer of ResNet34 after pruning on the ImageNet dataset while reducing the network FLOPs by 45\%.}
  \label{fig:figure6}
\end{figure}

The distribution of pruned filters among four different filter selection criteria is illustrated in~\figref{figure5}. For the ResNet50 model, most of the filters were pruned using magnitude-based criteria, while for the ResNet34 model, pruning was performed using $l_1$-norm, $l_2$-norm, and Euclidean similarity-based filter selection criteria. From ~\figref{figure4} and~\figref{figure5}, it can be observed that there is no consistency among models for different filter selection criteria, and it cannot be generalized. This demonstrates that our method is an appropriate technique for pruning filter selection as it dynamically selects the pruning criteria based on the actual filter weight distributions for each model. But, as we increase the number of filter importance criteria in a pool of criteria, the computation cost of the algorithm will also increase, therefore it is required to carefully select as minimum criteria as possible in the pool. Finally, ~\figref{figure6} shows the pruned filters and retained filters among the total number of filters in each convolutional layer of the ResNet34 model in the ImageNet dataset while reducing the FLOPs by $45\%$. We can see a similar effect as seen in~\figref{figure4} in terms of the pruning ratio in each layer.

\subsection{Ablation Studies}

\subsubsection{When to Prune?}

One of the critical considerations in our pruning algorithm is determining the appropriate time to perform pruning during training. If the network is pruned too soon, it may not have been trained adequately to accurately analyze filter weights for pruning, leading to suboptimal results. Conversely, if pruning is performed too late, there may not be enough remaining training epochs to recover the accuracy of the pruned network. To address this issue, we perform experiments to identify the optimal time for pruning. 

\begin{figure}[H]
  \centering
  \includegraphics[width=1.0\linewidth]{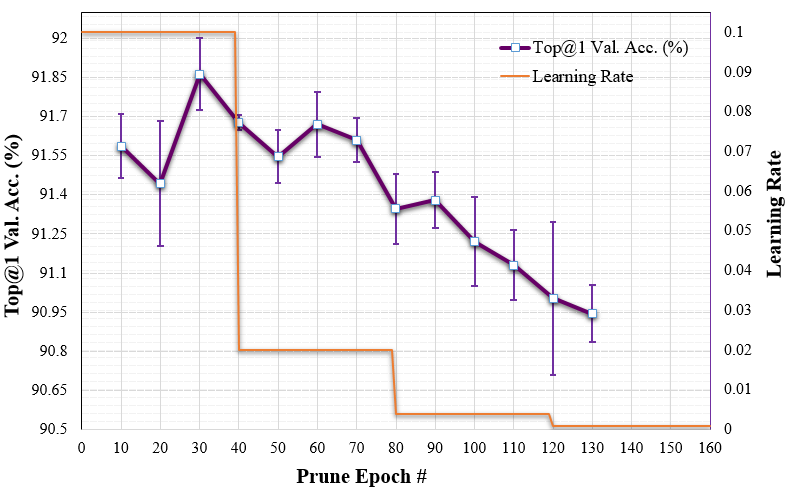}
  \caption{Top-1 validation accuracy after pruning ResNet20 model for CIFAR-10 dataset at different network training epochs while reducing the network FLOPs by 30\%.}
  \label{fig:figure7}
\end{figure}

As demonstrated in~\figref{figure7}, we prune the ResNet20 model on the CIFAR-10 dataset at different training epochs and then continue normal training until we reach the maximum training epochs. The results indicate that pruning at early epochs produces the best outcomes, with accuracy decreasing gradually as the pruning is delayed. Based on these findings, we conclude that pruning just prior to the decay of the initial learning rate provides the best results in terms of achieving the accuracy of the pruned network comparable to that of the original networks. The one obvious reason for consistently decreasing accuracy if the pruning is delayed during training under the same training cost is that there will not be left enough training epochs to recover the accuracy. Therefore we can argue that although we delayed the pruning if we increase the training cost for pruned network there will be better accuracy gain. But, our goal is to find the optimal pruning time during training while keeping the same overall training cost, therefore pruning in early epochs as demonstrated by~\figref{figure7} produces better results for a given fixed training cost.

\subsubsection{Pool of Importance Estimation Criteria}

In this study, we employed four importance estimation criteria, each with distinct properties.~\tabref{tbl4} illustrates the results obtained by utilizing magnitude criteria, similarity criteria, and a combination of both. As the cardinality of the set of criteria increases, the pruning complexity also increases linearly. However, this also means that the algorithm has a greater search space for identifying the set of pruning filters that have the least impact on overall network loss. Our experiments showed that employing all four criteria yielded superior results compared to using a subset of them. Thus, if pruning cost is not a significant concern, we suggest utilizing even more importance estimation criteria. However, if the pruning cost is limited, it is essential to select an optimal subset of criteria from the available pool.

\begin{table} [h]
\caption{Validation accuracy of ResNet32 model on CIFAR-10 dataset while reducing the FLOPs by $53\%$ with a different subset of filter ranking criteria.}
\label{tab:tbl4}
  \begin{center}
    {\small{
\begin{tabular}{lccc}
\toprule
{Criteria (C)} & \{$l_1$-norm, $l_2$-norm\} & \{eucl, cosine\} & \{$l_1$-norm, $l_2$-norm, eucl, cosine\}\\
\midrule
    Top-1 Acc. & {92.35 $\pm$ 0.14} & {92.59 $\pm$ 0.15} & \textbf{92.64 $\pm$ 0.09}\\
\bottomrule
\end{tabular}
}}
\end{center}
\end{table}

\subsubsection{Layer-wise Maximum Pruning Limit}

One of the critical hyperparameters that require setting in our pruning algorithm is the maximum allowed pruning rate for a convolution layer. This is a critical question for any pruning algorithm as if not properly defined, certain layers could be excessively pruned, making it difficult to recover the accuracy later. To address this, we conducted experiments on the ResNet20 model using the CIFAR-10 dataset, experimenting with various pruning limits for a layer while reducing the FLOPs by $50\%$.~\tabref{tbl5} shows the average results from three experiments. From the table, it is evident that setting the maximum pruning limit within the range of $0.55 \leq {R}^{(l)}{max} \leq 0.75$ does not consistently increase or decrease accuracy, at least for the ResNet20 model on the CIFAR-10 dataset. As a result, in our experiments, we set layer-wise pruning limits within the range of $0.55 \leq {R}^{(l)}{max} \leq 0.75$, based on the target FLOPs reduction rate. If a large reduction rate is desired, we use a larger ${R}^{(l)}_{max}$, and vice versa.

\begin{table} [h]
\caption{Validation accuracy of ResNet20 model while reducing the FLOPs by $50\%$ with varying ${R}^{(l)}_{max}$ on the CIFAR-10 dataset.}
\label{tab:tbl5}
  \begin{center}
    {\small{
\begin{tabular}{lccccc}
\toprule
{${R}^{(l)}_{max}$} & {0.55} & {0.60} & {0.65} & {0.70} & \textbf{0.75}\\
\midrule
    Top-1 Acc. & 91.62 $\pm$ 0.14 & 91.44 $\pm$ 0.02 & 91.45 $\pm$ 0.13 & 91.46 $\pm$ 0.03 & \textbf{91.65 $\pm$ 0.24}\\
\bottomrule
\end{tabular}
}}
\end{center}
\end{table}

\subsubsection{Influence of Exploration Steps}

The layer-wise exploration step in the proposed pruning technique can be adjusted by modifying the value of $\mathcal{P}_s$ in equation~\eqnref{step_rate}. The size of the exploration step directly affects the algorithm's complexity and the accuracy of the pruned network. Reducing the exploration step will increase the accuracy but also increase the complexity, and vice versa. We conducted experiments with different exploration step sizes and found that a value of $\mathcal{P}_s=0.01$ resulted in a reasonable trade-off between accuracy and complexity, as shown in~\tabref{tbl6}. This means that in each iteration of our pruning method, we can reduce the network's floating-point operations (FLOPs) by approximately $1\%$.

\begin{table} [h]
\caption{Validation accuracy of ResNet20 model while reducing the FLOPs by $50\%$ with varying $\mathcal{P}_s$ on the CIFAR-10 dataset.}
\label{tab:tbl6}
  \begin{center}
    {\small{
\begin{tabular}{lcccccc}
\toprule
{$\mathcal{P}_s$} & {0.005} & {0.01} & {0.015} & {0.02} & {0.025} & {0.03}\\
\midrule
    Top-1 Acc. & {91.47 $\pm$ 0.21}  & 91.65 $\pm$ 0.24 & \textbf{91.66 $\pm$ 0.09} & {91.39 $\pm$ 0.28} & {91.45 $\pm$ 0.22} & {91.42 $\pm$ 0.24}\\
\bottomrule
\end{tabular}
}}
\end{center}
\end{table}

\section{Conclusion}
\label{sec:conclusion}

In this work, we introduce a highly effective data-driven variable rate filter pruning method, named LAASP, for accelerating deep CNNs. Our experimental results, conducted on both small and large-scale datasets with different CNN architectures, demonstrate that LAASP outperforms state-of-the-art pruning methods. Our approach consists of iteratively identifying the optimal filter selection criteria and pruning layer, utilizing a small subset of the training data, to minimize the network loss. The partially trained network is utilized for pruning, and upon completion of this iterative pruning process, the resulting pruned network undergoes further training for the remaining number of original training epochs. Several ablations studies are also conducted to determine optimal hyperparameters used by the proposed pruning algorithm. In conclusion, our method requires only a minimal increase in the number of training epochs compared to the original network training without pruning. We also find that the automatic determination of pruning criteria, based on the distribution of filter weights among layers and networks, leads to the best validation results. 

In future studies, we aim to evaluate the effectiveness of our method on more advanced CNNs and applications, such as object detection and segmentation. Additionally, within the same pruning framework, we will examine more data-dependent and also data-independent pruning techniques, with a focus on reducing the cost associated with pruning. It would also be intriguing to automate the process of determining the appropriate early pruning time during training and introduce sparsity learning to evaluate the entire set of pruning parameters for their significance in the network. 

\bibliography{egbib}

\end{document}